\documentclass[letterpaper]{article} 
\pdfoutput=1
\usepackage{aaai25_modified}  
\usepackage{times}  
\usepackage{helvet}  
\usepackage{courier}  
\usepackage[hyphens]{url}  
\usepackage{graphicx} 
\usepackage{natbib}  
\usepackage{caption} 
\usepackage[T1]{fontenc}
\usepackage{amsmath}
\usepackage{header}
\usepackage{tikz}
\usepackage{circuitikz}
\usepackage{hyperref}

\usepackage{tcolorbox}
\usepackage{multicol}
\usepackage[acronym]{glossaries}
\newacronym{llm}{LLM}{large language model}
\newacronym{mdh}{MDHyp}{Marginalized Distributional Hypotheses}
\usepackage{mdframed}
\usepackage{newfloat}
\usepackage{stfloats}
\usepackage{nameref}
\usepackage{lipsum}
\usepackage{varioref}
\usepackage{cleveref}
\usepackage{dsfont}
\usepackage{cleveref}
\usepackage{mathtools}

\usepackage{listings}
\usepackage{todonotes}

\usepackage{tabularx} 
\usepackage{array}    

\DeclareCaptionStyle{ruled}{labelfont=normalfont,labelsep=colon,strut=off}

\usepackage{xcolor}

\newenvironment{redenv}{
    \color{red}
}{
    \ignorespacesafterend
}

\newenvironment{orgenv}{
    \color{orange}
}{
    \ignorespacesafterend
}

\newenvironment{blueenv}{
    \color{blue}
}{
    \ignorespacesafterend
}

\newenvironment{purpleenv}{
    \color{black}
}{
    \ignorespacesafterend
}

\newenvironment{oliveenv}{
    \color{olive}
}{
    \ignorespacesafterend
}

\newenvironment{tabenv}
   {\list{}{}%
    \item\relax}
   {\endlist}

\title{Can LLMs Reliably Simulate Human Learner Actions?\\ A Simulation Authoring Framework for Open-Ended Learning Environments}

\author {
    Amogh Mannekote\textsuperscript{\rm 1},
    Adam Davies\textsuperscript{\rm 2},
    Jina Kang\textsuperscript{\rm 2},
    Kristy Elizabeth Boyer\textsuperscript{\rm 1}
}
\affiliations {
    \textsuperscript{\rm 1}University of Florida\\
    \textsuperscript{\rm 2}University of Illinois Urbana-Champaign\\
    amogh.mannekote@ufl.edu, adavies4@illinois.edu, jinakang@illinois.edu, keboyer@ufl.edu
}

\usepackage{booktabs}
\usepackage{multirow}

\usepackage{listings}
\begin{document}

\maketitle

\begin{abstract}
    Simulating learner actions helps stress-test open-ended interactive learning environments and prototype new adaptations before deployment. While recent studies show the promise of using large language models (LLMs) for simulating human behavior, such approaches have not gone beyond rudimentary proof-of-concept stages due to key limitations. First, LLMs are highly sensitive to minor prompt variations, raising doubts about their ability to generalize to new scenarios without extensive prompt engineering. Moreover, apparently successful outcomes can often be unreliable, either because domain experts unintentionally guide LLMs to produce expected results, leading to self-fulfilling prophecies; or because the LLM has encountered highly similar scenarios in its training data, meaning that models may not be \emph{simulating} behavior so much as \emph{regurgitating} memorized content. To address these challenges, we propose \textsc{Hyp-Mix}, a simulation authoring framework that allows experts to develop and evaluate simulations by combining testable hypotheses about learner behavior. Testing this framework in a physics learning environment, we found that GPT-4 Turbo maintains calibrated behavior even as the underlying learner model changes, providing the first evidence that LLMs can be used to simulate realistic behaviors in open-ended interactive learning environments, a necessary prerequisite for useful LLM behavioral simulation.
\end{abstract}

\section{Introduction}\label{sec:intro}

Open-ended interactive learning environments offer unique educational value by providing tailored and dynamic spaces where learners can explore, experiment, and construct knowledge—capabilities \cite{renkl2007interactive, hannafin2013open, land2012theoretical}.
However, developing these environments is challenging. It requires not only the creation of pedagogical content but also mechanisms to adapt learning experiences for learners with varying knowledge levels and psychological characteristics for very large state spaces due to the relatively open-ended nature of the environments \citep{kim2012role, hannafin2014student, 2024.EDM-short-papers.51}. This complexity necessitates an iterative process in which theoretical best practices are continuously balanced with practical demands \cite{sandoval2014conjecture}.

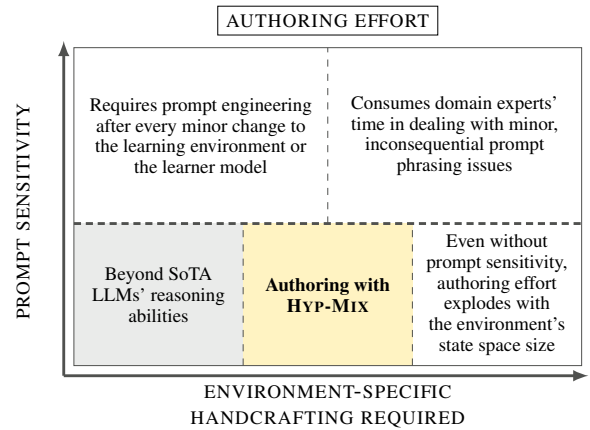
\begin{figure}[!t]
\begin{center}
\scalebox{0.9}{
\begin{tikzpicture}

\begin{scope}[every node/.style={align=center, scale=0.8}]

\draw[fill=lgrey,draw=none] (1,0.5) rectangle (3.5,2.6); 
\node at (2.25,1.55) {Beyond SoTA\\LLMs' reasoning\\abilities};

\draw[fill=lyellow,draw=none] (3.5,0.5) rectangle (6,2.6); 
\node at (4.75,1.55) {\textbf{Authoring with}\\\textbf{\textsc{Hyp-Mix}}};

\draw[draw=none] (6,0.5) rectangle (8.5,2.6); 
\node at (7.25,1.55) {Even without\\prompt sensitivity,\\authoring effort\\explodes with\\the environment's\\ state space size};

\draw[draw=none] (1,2.6) rectangle (4.75,5.2); 
\node at (2.875,3.9) {Requires prompt engineering\\ after every minor change to \\ the learning environment or\\ the learner model};

\draw[draw=none] (4.75,2.6) rectangle (8.5,5.2);
\node at (6.625,3.9) {Consumes domain experts'\\ time in dealing with minor,\\ inconsequential prompt\\ phrasing issues};

\end{scope}

\node[draw] at (4.75,5.6) {\textsc{authoring effort}};

\draw[draw=dgrey,thin] (1,0.5) rectangle (8.5,5.2);
\draw[draw=dgrey,thin,dashed] (3.5,0.5) -- (3.5,2.6);
\draw[draw=dgrey,thin,dashed] (6,0.5) -- (6,2.6);
\draw[draw=dgrey,thin,dashed] (4.75,2.6) -- (4.75,5.2);

\draw[draw=dgrey,very thick,densely dashed] (1,2.6) -- (8.5,2.6);

\draw[draw=dgrey, very thick, ->,>=latex] (0.85,0.35) -- (0.85,5.3);
\draw[draw=dgrey, very thick, ->,>=latex] (0.85,0.35) -- (8.6,0.35);

\node[align=center] at (4.75,0) {\\\textsc{environment-specific}\\\textsc{handcrafting required}};

\node[rotate=90,align=center] at (0.25,2.78) {\textsc{prompt sensitivity}};

\end{tikzpicture}
}
\end{center}
\caption{
We characterize the effort involved in authoring LLM-based simulations of learner behavior as a function of two key attributes of the simulation authoring process: 1) prompt sensitivity and 2) the extent of environment-specific handcrafting required during development. High prompt sensitivity necessitates excessive editing for minor phrasing changes, thus consuming valuable expert time. On the other hand, the need for environment-specific handcrafting arises when an LLM struggles to generalize across learning environments, impeding rapid iteration. The proposed approach of mixing-and-matching expert-hypotheses to define simulation behavior offers a promising balance, enabling authors to impose necessary constraints while leveraging the advantages of state-of-the-art knowledge and reasoning capabilities of LLMs for ``filling in the gaps.''
}
\label{fig:challenge}
\end{figure}

Simulations of learner behavior have been instrumental in streamlining the process of developing intelligent systems for education \citep{koedinger_methods_2015, matsuda2015teaching}. By allowing developers to rigorously test features before full deployment, these simulations reduce reliance on resource-intensive pilot testing in real-world classrooms \citep{kaser_simulated_2023}. They enable developers to identify software issues and evaluate design choices early, later fine-tuning the environment to meet learner needs. However, developing simulations during the cold-start phase is challenging due to the lack of real-learner data in new environments. This scarcity prevents purely data-driven approaches, requiring reliance on log data from similar studies, predictions from learning science theories, instructor experience, and expert intuition \citep{holstein2019co}. Without action logs from the target demographic, these sources provide the best alternative for accurate simulations.

Combining these alternative information sources to craft realistic simulations of learner actions requires a balanced integration of expert knowledge and automated reasoning. Fully handcrafted, rule-based simulations offer fine-grained control but become impractical as complexity increases, while purely automatic systems may miss critical nuances \citep{wang2024survey}. LLM prompting may potentially strike an ideal balance, using rich natural language to specify behavior while leveraging the LLM's reasoning capabilities. This approach holds the potential for flexible, fine-tuned simulations that effectively bridge the gap between manual control and automation.

Promisingly, there has been a recent surge in studies that suggest that LLMs, with their extensive world knowledge and reasoning capabilities, can accurately predict human responses to both natural language descriptions of hypothetical situations and actual experimental setups taken from academic disciplines like psychology and behavioral economics \citep{aher_using_2023}. However, such claims must be approached with caution. We identify three reasons to be skeptical of simulations based on \gls*{llm} prompting reliably generalizing to new situations.
\begin{enumerate}
    \item LLMs are known to be highly sensitive to small, inconsequential changes to the prompt wording (prompt sensitivity) \citep{sclar2023quantifying, loya2023exploring}. As a result, a simulation that works in one context might fail with changes to either the description of the learning environment (corresponding to, say, a new feature that the developers are planning to add to the environment) or the learner model (corresponding to refinements in the expert's understanding of how learners behave).
    \item LLMs, trained on vast web data, may rely on memorization rather than genuine reasoning, limiting their ability to generalize \citep{sainz-etal-2023-nlp}. 
    \item There is no disciplined method to prevent prompt engineers from unconsciously shaping prompts to elicit expected answers, raising concerns of a Clever Hans\footnote{The term originates from Hans, a horse in early 20th century Germany, who seemed to perform arithmetic by tapping his hoof. It was later found he was responding to subtle cues from his trainer or the audience.}-like setup, where human cues influence the outcome \citep{kambhampati_can_2024}.
\end{enumerate}

For the reasons stated above, the usefulness of LLMs for simulating learner actions beyond single proof-of-concept experiments has not yet been established.
To address this gap, we introduce a simulation authoring framework that serves the dual purposes of: 1) systematically evaluating whether an \gls*{llm}-based simulator can usefully generalize to new contexts (e.g., modifications of the original learning environment or the original learner model) without re-engineering the \gls*{llm} prompt; and 2) establishing a clear prompting workflow to avoid Clever-Hans-style biases, preventing overestimation of the \gls*{llm}'s capabilities.

A robust simulator must dynamically adapt to changes in the simulation context (learning environment or learner model) without extensive prompt recalibration. Once a prompt template is calibrated to specific learner behaviors, this calibration should generalize to new simulation contexts, maintaining consistent simulation behavior. 
This generalization is important for two reasons: (1) the exponential increase in experiment runs needed as state variables grow, and (2) the limited utility of LLM simulations that only predict behaviors when specifically calibrated, which fails to generate new insights and merely reproduces existing findings (Clever Hans effect).

Our main contribution with \textsc{Hyp-Mix} is a systematic simulation authoring framework\footnote{\href{https://github.com/msamogh/hypmix}{https://github.com/msamogh/hypmix}} for incorporating expert knowledge into \gls*{llm}-based simulations of learner actions. 
Our hypothesis-based framework presents a well-defined, statistical notion of what it means for the simulation to be robust and generalizable to new simulation scenarios. 
Using our framework, we find that GPT-4 Turbo is capable of maintaining prompt calibration under changes to the learner model, indicating that it may already be feasible to simulate realistic learner behaviors in learning environments using frontier LLMs.

\section{Related Work}

    \paragraph{Simulated Learner Behavior for Authoring Educational Technologies.}
    Simulated learners streamline the authoring of intelligent tutoring systems (ITSs), which often require over 100 hours of work per instructional hour \citep{blessing2008evaluating}. Tools like SimStudent \citep{matsuda2015teaching} simulate learner behavior to aid in ITS development via interactive tutoring. However, compared to ITSs, open-ended interactive learning environments typically involve more states due to their open-ended and exploratory nature and a greater emphasis on scaffolding the affective aspects of learning \citep{rieber1996seriously}. While \citet{christensen2011simschool} simulate psychological aspects of learners, their method is handcrafted, highly context-specific, and therefore, would not scale well to complex interactive environments. To our knowledge, our work is the first to apply learner behavior simulations to these environments. Additionally, \citet{kaser_simulated_2023} identify a widespread lack of validation in simulated learner research, which we address in the \textsc{Hyp-Mix} framework by centering on falsifiable hypotheses for both authoring and evaluation. Our approach is also in line with \citet{redeem}, who focus on group-level behavior specification, similar to our use of distributional hypotheses.

    \paragraph{Simulating Human Behavior with LLMs}
    Several recent works explore the ability of \glspl*{llm} to simulate human behaviors across various contexts, including social platform design \cite{park2022social}, market research \citep{brand2023using}, and experimental economics \citep{Gui2023TheCO}.
    LLMs have also been shown to reflect human-like cognitive biases in reasoning tasks \citep{dasgupta2022contenteffects, ozeki2024reasoning}.
    Most related to our work are studies that analyze \glspl*{llm} agents' consistency with provided personality traits \cite{frisch_llm_2024,jiang2024personallm} or character profiles \cite{xiao2023far}.
    However, in contrast to these works, we evaluate agent consistency using simple hypotheses specifying the statistical relationship between values of agent (learner) characteristics and behaviors, alleviating the need for fine-grained annotation of individual responses;
    and further consider how these simulated behaviors change in response to changes in the simulation context.

    \paragraph{Prompt Sensitivity and Prompt Calibration.}
    Experiments using LLMs rely heavily on natural language prompts to define personas, situations, and tasks, but LLMs are highly sensitive to slight variations in prompt text, making this a critical issue for research \citep{mohammadi_wait_2024}. \citet{loya-etal-2023-exploring} find that ChatGPT exhibits sensitivity to prompt phrasing for decision-making tasks such as ours.
    In response, various prompt calibration approaches have emerged, particularly focusing on reducing the \glspl*{llm}' sensitivity to the order of in-context examples \citep{lu-etal-2022-fantastically, zhao_calibrate_2021}. 
    In contrast to this family of work that focuses on reducing variance between different templates, in this work we our goal is to test the consistency of LLM behaviors across different simulation contexts.

\section{The \textsc{Hyp-Mix} Framework}

The \textsc{Hyp-Mix} framework is designed to create and evaluate realistic and scalable simulations of learner behavior by translating theoretical constructs into concrete, testable predictions. The unit of authoring and evaluation in this framework is a \gls*{mdh}. These are called ``marginal'' because they focus on one learner characteristic at a time, while ``marginalizing'' over all other variables. This is essential because, while it is straightforward to reason about a single characteristic, jointly considering multiple characteristics can quickly become difficult. For instance, an MDHyp might predict that low persistence leads to a higher probability of task-abandonment, focusing specifically on persistence while accounting for other variables in the background.
The rest of this section details the motivation and implementation of \glspl*{mdh}, along with its integration with \gls*{llm} prompting.

\subsection{\glspl*{mdh} for Simulation Evaluation}

A common method for validating simulated agents involves presenting the generated behaviors to human crowdworkers, who then rate the realism of these behaviors either over a quantitative scale or according to a qualitative rubric \citep{park_generative_2023,jiang2024personallm}. While this approach has been widely adopted in recent studies, particularly with the proliferation of crowdsourcing platforms, it is fundamentally limited and ill-suited for evaluating simulated learner actions in complex, iterative experiments, for several reasons:
\begin{enumerate}
    \item \textbf{Cost Constraints:}  Crowdsourcing becomes prohibitively expensive in iterative studies, particularly those requiring extensive experimentation.
    \item \textbf{State Space Explosion:} As the complexity of the environment and the number of learner characteristics increase, the task of collecting annotations for every possible combination becomes infeasible.
    \item \textbf{Demographic Mismatch:} The typical crowdworker populace does not include individuals deeply involved in education, such as researchers or educators \citep{huff2015these}. As a result, they are generally not equipped to accurately assess the realism of behaviors exhibited by young learners with specific characteristics.
    \item \textbf{Inherent Noise in Learners' Actions:} The stochastic nature of interactions within learning environments introduces significant noise into the evaluation process. Even with a sophisticated model of a learner, it is nearly impossible to predict with certainty how they will behave in a given situation, making deterministic point estimates unreliable.
\end{enumerate}

We propose using \glspl*{mdh} to evaluate learner behavior simulations at a distributional level, drawing from prior studies or instructor experience. An \gls*{mdh} is a natural language statement that describes a relationship between a learner's characteristic and their probability of taking certain actions (e.g., ``\textit{a more persistent learner is less likely to abandon the task as more time passes}''). This relationship can be tested by analyzing the distribution of outcomes from multiple simulation runs across different environment states.

\subsection{\glspl*{mdh} for Simulation Authoring}
 
The central thesis of the \textsc{Hyp-Mix} framework is that \glspl*{mdh} serve not only as useful tools for \textit{evaluating} an existing simulation, but also as powerful building blocks for \textit{expert-authoring} \gls*{llm}-based simulations of learner behavior.

\paragraph{Achieving Mix-and-Match Simulation Authoring with \glspl*{mdh}}
For \glspl*{mdh} to be effective in prompt-based simulation authoring, the \gls*{llm} must demonstrate compositional generalization \citep{compositional-generalization}. We need \glspl*{mdh} to function as modular elements that can be easily added, edited, removed, swapped, and combined to shape the \gls*{llm}'s outputs. Similar to \textsc{Skill-Mix} \cite{yu2023skillmix}, which tests \glspl*{llm}' ability to combine literary and logical devices to generate free-form text, \textsc{Hyp-Mix} tests \glspl*{llm}' ability to combine calibrated expert-hypotheses to simulate learner actions in a ``calibrate once, use forever'' fashion. Achieving this, of course, is challenging due to \glspl*{llm}' sensitivity to prompt phrasing and requires empirical validation.

\paragraph{Existing Notions of ``Calibration''} The term ``calibration'' carries different definitions across disciplines. In statistics and machine learning, calibration refers to the alignment between a model's predicted probabilities and the actual observed frequencies of outcomes, ensuring that predictions accurately reflect real-world occurrences over time \citep{bella2010calibration}. 
In the context of physical measurement devices, calibration ensures that a measurement device's accuracy is consistent. This process involves aligning the device with a known standard to maintain reliable accuracy across future measurements \citep{castrup1994metrology}.
The \textsc{Hyp-Mix} notion of calibration combines the two: we want the predicted action probabilities from the \gls*{llm} to align with the \gls*{mdh} (analogous to the statistical notion) and also to hold this calibration across different hypotheses and changes in the underlying learner model (analogous to the metrological notion).

\paragraph{Holding Calibration}
Building on this integrated definition, the ability of an \gls*{llm} to hold calibration of a prompt template across simulation contexts is critical for minimizing the labor-intensive re-engineering of prompt templates after each modification to the simulation model. By grouping hypotheses into \textit{hypothesis classes} based on similar functional relationships and linking them to specific statistical tests, we ensure robust calibration, even as the simulation model undergoes modifications.

\paragraph{Hypothesis Classes} A hypothesis class defines a specific functional relationship that its member-hypotheses posit between independent variables (e.g., learner persona values, environment state variables) and a dependent variable (e.g., probability mass of specific learner actions). Formally, a hypothesis $H_i$ belongs to the hypothesis class $\mathcal{H}_\text{class}$ (denoted as $c(H_i) = \mathcal{H}_\text{class}$). Each hypothesis class is associated with a prompt template, $\hat{I}_\text{class}$, that its member-hypotheses instantiate by specifying slot values, and is linked to a specific success criterion $T_\text{class}$, typically expected to be the result of a statistical test (e.g., Chi-squared) designed to assess how well the LLM maintains consistency and accuracy when different instances of that class's characteristic relationship are tested (e.g., any relationship that can be expressed in natural language, such as linear, logarithmic, or piecewise continuous relationships).

\paragraph{Template Calibration} Finally, template calibration is the human-in-the-loop ``prompt engineering'' process that involves iteratively refining the prompt template associated with a hypothesis class until the \gls*{llm}'s output probabilities for specific actions align with one or more member-hypotheses that are used as validation. Successful calibration is achieved when statistical tests confirm this alignment. The calibrated prompt template should remain robust under changes in the learner model and across new member-hypotheses. \Cref{tab:hypotheses} shows the set of instantiated prompts corresponding to the hypothesis classes and hypotheses that we use in our illustrative experiments that are described later in the paper.

\section{Experiments}

We test the robustness and generalization capabilities of GPT-4 Turbo, a state-of-the-art \gls*{llm}, in the HoloOrbits environment by assessing how well it maintains calibration when the learner model is modified.

\subsection{Learning Environment}

\begin{figure}[t!]
\centering
\includegraphics[scale=0.14]{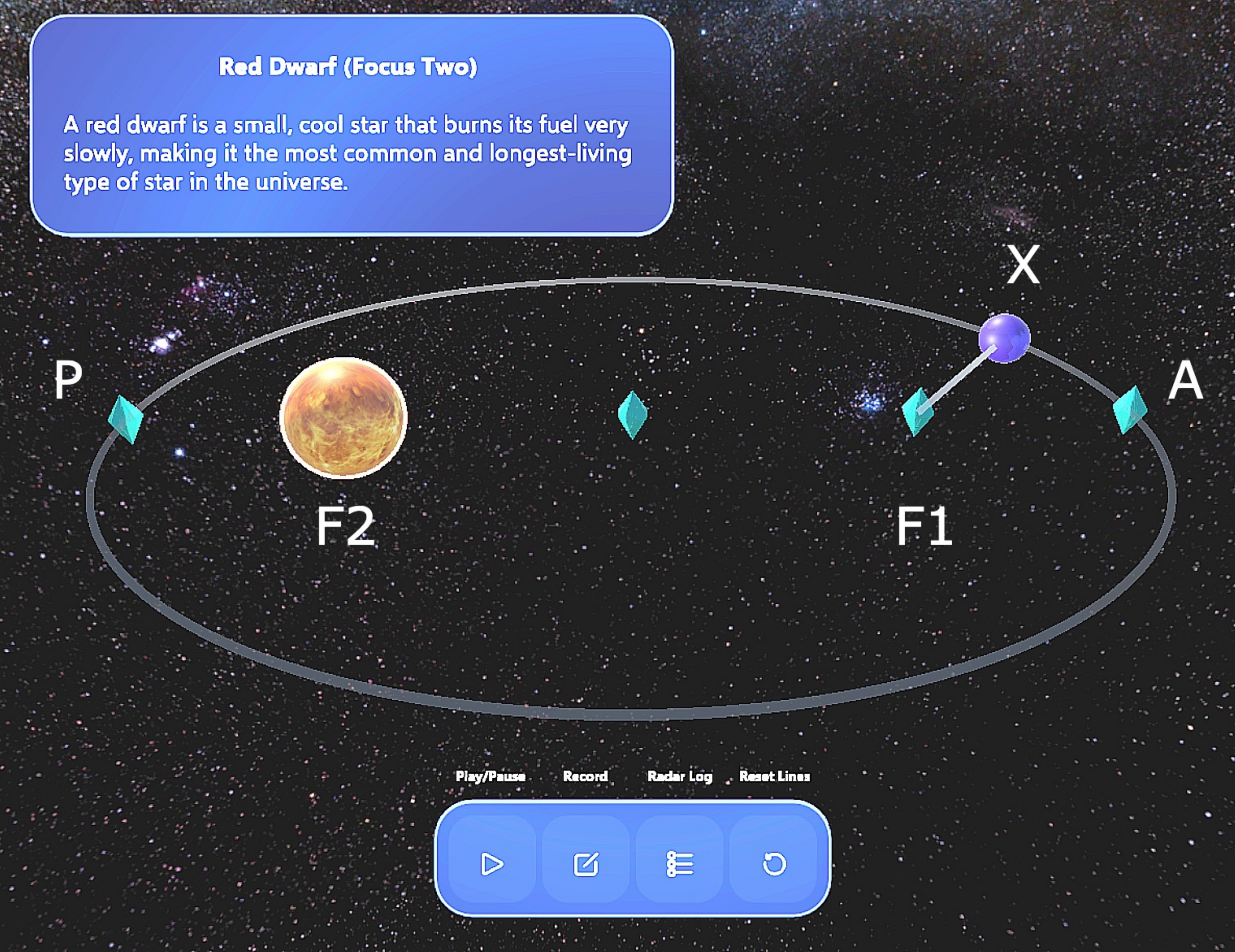}
\caption{A screenshot of the original HoloOrbits environment \citep{2024.EDM-short-papers.56} with the keypoints annotated.}
\label{fig:holoorbits}
\end{figure}

We situate our experiments within HoloOrbits \citep{2024.EDM-short-papers.56}, an open-ended interactive learning environment designed for teaching Kepler's Laws that we use for our experiments (see \Cref{fig:holoorbits}). We selected HoloOrbits due to its small, well-defined state and action spaces, which make it ideal for preliminary experiments.
Since our experiments involve a text-based \gls*{llm}, we only need a textual description of the learning environment to feed into the model as a natural language prompt. An unintended advantage of this approach is that it allows us to describe learning environments not yet implemented in software.

\subsubsection{Learning Task}
We particularly focus on the learner's task to verify if a given planetary system adheres to Kepler's First Law by submitting three equal arithmetic expressions. These expressions can use any combination of distance measurements between the following points: aphelion (\textsc{\textbf{a}}), perihelion (\textsc{\textbf{p}}), focus 1 (\textsc{\textbf{f1}}), focus 2 (\textsc{\textbf{f2}}), and a fixed point on the orbit (\textsc{\textbf{x}}). The correct solution involves submitting the following measurements: \textsc{\textbf{(f1-a + f2-a)}}, \textsc{\textbf{(f1-p + f2-p)}}, and \textsc{\textbf{(f1-x + f2-x)}}.

\subsubsection{State Representation}
For our experiments, we define a minimal state representation with ten boolean variables indicating whether the learner has measured the distances between each pair of points. Additionally, we include two integer variables to track the number of submission attempts and the time elapsed since the session began, respectively.

\subsubsection{Action Space}

The learner can perform measurements between any pairs of key points in the planetary system, with specific actions such as \textsc{\textbf{measure-f1-x}} to measure the distance between Focus 1 (\textsc{\textbf{f1}}) and a fixed point on the orbit (\textsc{\textbf{x}}), \textsc{\textbf{measure-a-f1}} to measure the distance between Aphelion (\textsc{\textbf{a}}) and Focus 1 (\textsc{\textbf{F1}}), or \textsc{\textbf{measure-a-p}} to measure the distance between Aphelion (\textsc{\textbf{a}}) and Perihelion (\textsc{\textbf{p}}).
In addition to these measurement actions, the learner can submit solutions using \textsc{\textbf{submit(x, y, z)}}, where \textsc{\textbf{x}}, \textsc{\textbf{y}}, and \textsc{\textbf{z}} represent arithmetic expressions involving the measured distances. The goal is for all three expressions to be equal. The learner also has the option to \textsc{\textbf{exit}} at any time.

\begin{table}[t!]
\scalebox{0.77}{
\begin{tabular}{@{}l p{8cm}@{}}
\toprule
\textbf{\begin{tabular}[c]{@{}l@{}}Hypothesis\\ ($H_i \in H_{c(H_i)}$)\end{tabular}} &
  \textbf{\begin{tabular}[c]{@{}l@{}}Initial, Uncalibrated Prompt Template\\$\hat{I}_{c(H_i)}(H_i)$\end{tabular}} \\ \midrule
\begin{tabular}[t]{@{}l@{}} $H_{G1} \in \mathcal{H}_\text{mono}$ \ \end{tabular}&
A learner with a higher \color{blue}\textbf{geometry proficiency}\color{black}\ is more likely to \color{blue}\textbf{make productive measurements} \color{black}(i.e., \color{blue}\textbf{those that measure distances between pairs of points in the planetary system that are potentially useful to verify if the orbit is elliptical}\color{black}). To \color{blue}\textbf{make productive measurements}\color{black}\ is to make one of the following actions: \color{blue}\textbf{\textsc{measure-f1-x}, \textsc{measure-f2-x}, \textsc{measure-f1-p}, \textsc{measure-f2-p}, \textsc{measure-a-f1}, \textsc{measure-a-f2}}. \\\\
\begin{tabular}[t]{@{}l@{}} $H_{P1} \in \mathcal{H}_\text{mono}$ \end{tabular} &
A learner with a higher \color{blue}\textbf{persistence}\color{black}\ is \color{blue}\textbf{less}\color{black}\ likely to \color{blue}\textbf{abandon the task as the number of measurements increases}\color{black}\ (i.e., \color{blue}\textbf{to prematurely exit the session before submitting the right solution}\color{black}). To \color{blue}\textbf{abandon the task as the number of measurements increases}\color{black}\ is to make one of the following actions: \color{blue}\textbf{\textsc{exit}}\color{black}. \\\\
\begin{tabular}[t]{@{}l@{}} $H_{P2} \in \mathcal{H}_\text{mono}$ \end{tabular} &
A learner with a higher \color{blue}\textbf{persistence}\color{black}\ is \color{blue}\textbf{less}\color{black}\ likely to \color{blue}\textbf{abandon the task as the time elapsed increases}\color{black}\ (i.e., \color{blue}\textbf{to prematurely exit the session before submitting the right solution}\color{black}). To \color{blue}\textbf{abandon the task as the time elapsed increases}\color{black}\ is to make one of the following actions: \color{blue}\textbf{\textsc{exit}}\color{black}. \\\\
\begin{tabular}[t]{@{}l@{}} $H_{G2} \in \mathcal{H}_\text{uniform}$ \end{tabular} &
As learners get closer and closer to the \color{blue}\textbf{lower}\color{black}\ end of the \color{blue}\textbf{geometry proficiency}\color{black}\ spectrum (value of 1), they are equally likely to perform the following actions. In other words, such a learner exhibits a uniform distribution over these actions: \color{blue}\textbf{\textsc{<all measurement actions>}}\color{black}. \\ \bottomrule
\end{tabular}
}
\caption{Illustrative hypotheses used in our learner modeling experiments. The regular text represents the template for each hypothesis class, with \color{blue}\textbf{blue}\color{black}\ text indicating specific slot values filled by each hypothesis. The Appendix provides the updated, calibrated prompt templates.\label{tab:hypotheses}}
\end{table}

\subsection{Learner Model}

We represent each learner through a learner model $\mathcal{L} = (\mathcal{C}, \mathcal{V}, \mathcal{M})$. $C$ is the set of \textbf{learner characteristics} (e.g., geometry proficiency, persistence) being modeled. $\mathcal{V}$ is the mapping between each learner characteristic, $C_i \in \mathcal{C}$, to its corresponding \textbf{persona level}, $V_i$, of the current learner. Each $V_i \in \mathcal{V}$ is quantified on a numerical scale ($V_i \in [1, 10]$).
Finally, each learner characteristic $C_i$ is associated with a \textbf{learner characteristic model} $M_i \in \mathcal{M}$, which, in turn, comprises one or more \glspl*{mdh}.

\paragraph{Learner Characteristics}
For the HoloOrbits learning environment, we model learners using persistence (a psychological factor) and geometry proficiency (which reflects the learner's knowledge of the subject matter) with the following operating theoretical definitions:
\begin{itemize}
    \item \textbf{Persistence:} ``maintaining a sustained effort toward completion of a goal-directed task despite challenges or difficulties'' \citep{anderson2002assessment, hilton2012education}
    \item \textbf{Geometry Proficiency:} ``the ability to apply the knowledge of the properties of common shapes to solve problems'' \citep{educsci13070682}
\end{itemize}

\subsubsection{Design of the \gls*{llm} Prompt for the Simulation}

The simulation prompt fed to the \gls*{llm}, $\hat{I}_\text{sim}$, consists of introductory instructions, a description of the learning environment, current state, and the learner model (a graphic of the prompt template is shown in the Appendix). Furthermore, the prompt template for each learner characteristic model is a concatenation of prompt templates of the \glspl*{mdh} that make up the learner characteristic model. We also instruct the \gls*{llm} to perform Chain-of-Thought reasoning \citep{wei2022chain} before outputting the simulated action to strengthen the reasoning and provide the practitioners with a semblance of the intermediary steps used to arrive at the output, which can then be used to refine the simulation.

\begin{figure*}[t!]
    \centering
    \includegraphics[width=0.92\textwidth, page=1]{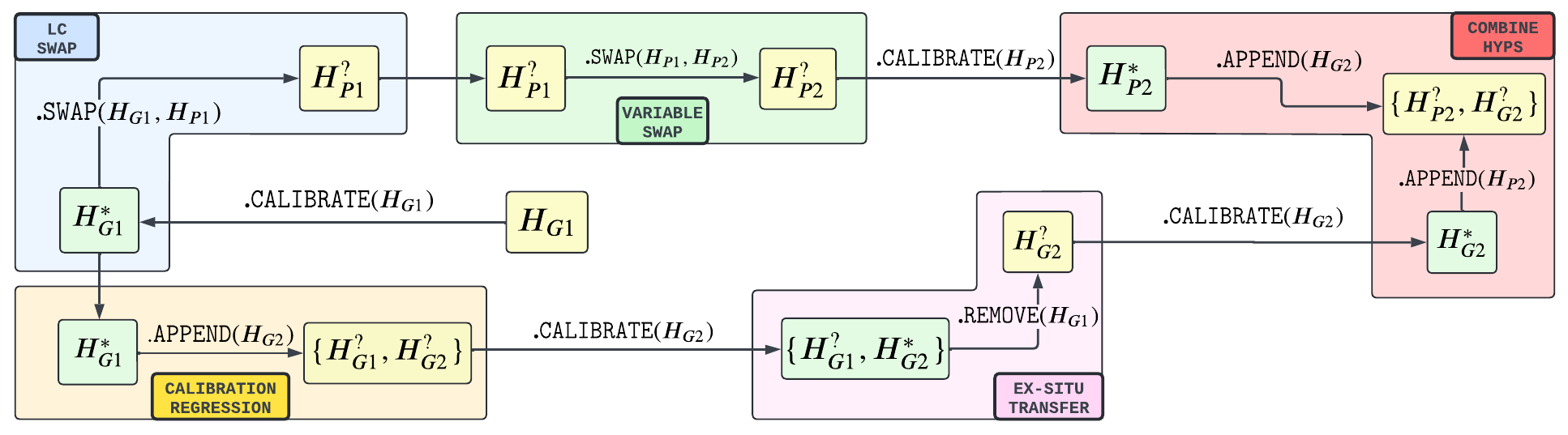}
        \caption{The Learner Model Edit Graph used in our experiments to evaluate \gls*{llm} robustness across five distinct edit operations to the learner model. Each node represents a ``snapshot'' of the learner model after specific edits by the developer. Inside each node, the \glspl*{mdh} comprising the learner model snapshot are listed. Green nodes indicate calibrated snapshots, while yellow nodes represent states untested for calibration. Each \gls*{mdh} in the learner model is annotated with a superscript: `?' for untested calibration status and `*' for confirmed calibration. \textbf{(1) Ex-Situ Transfer:} Tests if an \gls*{mdh} that is calibrated alongside other \glspl*{mdh} remains calibrated when tested alone. \textbf{(2) Combine Hypotheses:} Assesses if two separately calibrated hypotheses remain stable when combined. \textbf{(3) Variable Swap:} Involves swapping a single variable within a hypothesis. \textbf{(4) LC Swap:} Evaluates if a prompt template calibrated for one learner characteristic works for another in the same class. \textbf{(5) Calibration Regression:} Tests if a calibrated hypothesis remains stable when a new hypothesis is added to the model.}
    \label{fig:experiment-graph}
\end{figure*}

\subsubsection{Approximate Marginalization}
Testing an \gls*{mdh} by running the simulation over all value-assignments of state variables $\mathcal{S}$ requires an intractable number of LLM calls that grows exponentially with $|\mathcal{S}|$. To address this, we statistically approximate the state space by subsampling it.
This approach allows for manageable marginalization while controlling computational costs.

\subsection{Learner Model Edit Graph: A Case Study}

This section details a case study of how we developed a simulation of learner actions for the HoloOrbits environment leveraging the \textsc{Hyp-Mix} framework. The goal of this case study is to demonstrate how \textsc{Hyp-Mix} can be used to evaluate the compositional generalization capabilities of an \gls*{llm} (we used GPT-4 Turbo \citep{achiam2023gpt} in our experiments). We focus on five representative types of modifications (edit operations) to the learner model, reflecting the iterative process a developer might follow when constructing a learner model. Throughout the development process, we use the four \glspl*{mdh} listed in \Cref{tab:hypotheses}. These modifications are represented via a Learner Model Edit Graph (shown in \Cref{fig:experiment-graph}).

\begin{enumerate}
\item \textbf{Initial Hypotheses and Operationalization:} We initialize the learner model with two hypotheses: $H_{G1}$ and $H_{P1}$, both obtained by operationalizing the theoretical definitions of geometry proficiency and persistence respectively into \glspl*{mdh} (see \Cref{tab:hypotheses} for all hypotheses used).
Both $H_{G1}$ and $H_{P1}$ posit \textit{monotonic} relationships between variables. We grouped them under the hypothesis class $\mathcal{H}_\text{mono}$. We calibrated $\hat{I}_\text{mono}$ using $H_{G1}$ as the calibration reference hypothesis and tested for generalization on $H_{P1}$. We define the success criteria function for monotonic hypotheses, \( T_\text{mono} \) using the Spearman correlation coefficient \( \rho \) and its corresponding p-value $P_\rho$ as follows:
\begin{align}
T_\text{mono}(\rho, P_\rho) =
\begin{cases}
\textsc{true}, & \text{if } \rho > 0 \text{ and } P_\rho \leq 0.05 \\ 
   & \text{for a monotonically } \\
   & \text{increasing hypothesis} \\
\textsc{true}, & \text{if } \rho < 0 \text{ and } P_\rho \leq 0.05 \\
   & \text{for a monotonically } \\
   & \text{decreasing hypothesis} \\
\textsc{false}, & \text{otherwise}
\end{cases} 
\label{eq:mono} 
\end{align}
For $H_{G1}$, Spearman correlation is computed between the persona value for geometry proficiency and empirical probability of making a productive measurement.

\item \textbf{Variable Swap:}
After consulting with learning science experts, we determined that the ``Number of Submissions'' was a more suitable measure of ``challenge'' than ``Number of Minutes Elapsed.'' This led to a modification of the original \gls*{mdh} for persistence, resulting in a new hypothesis, $H_{P2}$.

\item \textbf{Append:}
During testing, we observed that learners with minimal Geometry Proficiency (1/10) were unexpectedly producing a high percentage ($\sim$80\%) of productive measurements, contrary to our expectation of a uniform distribution\footnote{We hypothesize this result to be a result of a more general phenomenon that \citet{aher_using_2023} refer to as ``hyper-accuracy distortion,'' where \glspl*{llm} struggle to feign ignorance about a topic to simulate human behavior.}. To address this, we introduced $H_{G2}$ and a new hypothesis class, $\mathcal{H}_\text{uniform},$ to explicitly model this behavior and refine the Geometry Proficiency model, better align it with our theoretical expectations.
The success criteria function \( T_{\text{uniform}} \) for the uniform distribution hypothesis is defined using the p-value of a Chi-squared test $P_{\chi^2}$ as follows:
\begin{align}
    T_{\text{uniform}}(P_{\chi^2}) = \textsc{true}\ \textrm{if}\ P_{\chi^2} > 0.05\ \textrm{else}\ \textsc{false} \label{eq:uniform}
\end{align}

\item \textbf{Combine Hypotheses:}
After calibrating the prompt templates for $H_{G2}$ and $H_{P2}$, we combined these \glspl*{mdh} into a unified learner model, completing the development process.
\end{enumerate}

\section{Results and Discussion}
The criterion for evaluating whether an \gls*{llm} can support flexible simulation authoring within the \textsc{Hyp-Mix} framework is that the calibration state of all hypothesis classes must remain intact after a learner model edit operation. Specifically, for each hypothesis $H_i$ in the learner model, the LLM outputs must continue to satisfy the success criterion function $T_{c(H_i)}$ without requiring any changes to the associated prompt template, $\hat{I}_{c(H_i)}$. We evaluate this by comparing the success criterion function outputs for all hypotheses in the learner model before (Pre-Op) and after (Post-Op) the edit operation. We use three different labelings of the action space to enhance the reliability of calibration and results. This approach helps rule out LLM sensitivity to minor label variations, such as \textsc{exit} vs. \textsc{quit}. Consequently, each operation in \Cref{tab:results} is represented by three rows (the action spaces are shown in the Appendix).

To better illustrate this process, consider the following example. According to the definition of $H_{G1}$, we expect the learner's empirical probability of making productive measurements between key points in the planetary system to increase monotonically with the learner's geometry proficiency level. Since $H_{G1} \in \mathcal{H}_\text{monotonic}$, we calibrate the template $\hat{I}_\text{monotonic}$ using $H_{G1}$ as the calibration reference hypothesis until the monotonicity test $T_\text{monotonic}$ is satisfied for $H_{G1}$.
Once $H_{G1}$ is successfully calibrated, we modify the learner model by introducing a new hypothesis, $H_{P1}$. We then reapply $T_\text{monotonic}$ to $H_{G1}$ (with $H_{P1}$ included in the learner model via the LLM prompt). We report whether the calibration state of $\hat{I}_\text{monotonic}$ is maintained or lost in our results (see \Cref{tab:results}). This procedure is repeated for all directed edges in the graph. 

We find that except for the \textsc{combine} operation where calibration was not maintained for two out of three action spaces,
GPT-4 Turbo succeeds in holding calibration through all the remaining four learner model edit operations (Table \ref{tab:results}).
This result indicates that GPT-4 Turbo was usually (in 16 of 18 cases) able to generalize to new learner models and characteristics without requiring re-calibration, which is important both for practical reasons (e.g., the cost associated with manual re-calibration) and more fundamental ones (e.g., generating novel insights rather than simply calibrating to reproduce existing findings).
The two instances where calibration was not maintained suggest that combining separately calibrated hypotheses might be more challenging yet the model demonstrates strong stability across other operations where multiple hypotheses are present.

While further experimental evidence is needed before we can generalize these claims across learning environments, learner characteristics, and even different \glspl*{llm}, the results from this illustrative experiment bode well for the use of \glspl*{mdh} as the unit of simulation-authoring with existing \gls*{llm} technology.
More broadly, balancing explicit and implicit authoring of agent simulations involves deciding which specific agent behaviors must be defined manually and which can be left for the LLM to handle automatically. In sensitive domains like education, a bias toward explicit authoring is prudent \citep{2024.EDM-posters.75}, as LLMs struggle with certain reasoning tasks \citep{huang2022towards, kambhampati_can_2024, kambhampati_llms_2024}. Our proposed \glspl*{mdh} and Learner Model Edit Graph abstractions offer a foundation for building benchmark datasets that evaluate LLM performance across different learner characteristics and multiple learning environments.

\begin{table*}[t!]
\centering
\includegraphics[scale=0.33]{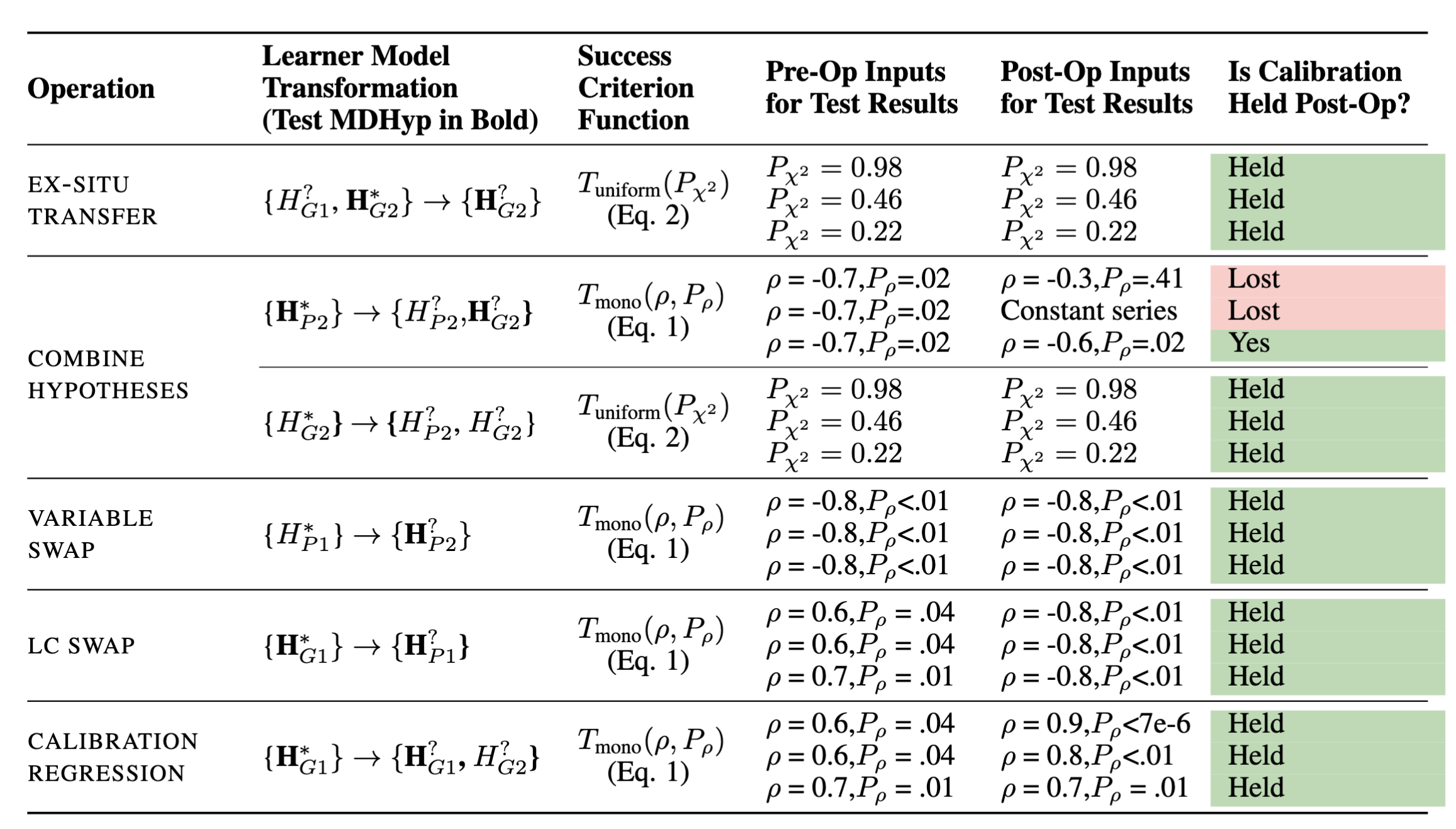}
\caption{Results of statistical tests evaluating the impact of different operations on the calibration state of hypotheses within the learner model. The table compares pre-operation (Pre-Op) and post-operation (Post-Op) results using Chi-squared and Spearman correlation tests, conducted across three different labelings of the action space for improved reliability. 
The operations include Ex-Situ Transfer, Combine Hypotheses, Variable Swap, LC Swap, and Calibration Regression. For each operation, the table provides the specific hypotheses tested, the applied statistical test, and the resulting p-values. Bolded hypotheses indicate those tested in both the pre- and post-op phases. Green shading denotes stable test results (holding calibration), red shading shows a total loss of calibration, and yellow shading indicates that the \gls*{mdh} is satisfied post-operation, though with some degradation in statistical significance.}
\label{tab:results}
\end{table*}

\section{Limitations and Future Work}

Our study focuses on two learner characteristics—geometry proficiency and persistence—within a single learning environment. While this scope is limited, it allows us to generate targeted insights and refine our methods, laying the groundwork for future expansion. Complex, non-linear interactions likely exist that are challenging to calibrate without real learner data \citep{klein-latucha_when_2024}. Future work should build on this foundation by exploring a broader range of environments and more intricate relationships among learner characteristics while assessing the robustness of LLMs to changes in both the learner model and the environment. Although \glspl*{mdh} offer scalable assessments, the real value of learner action simulations lies in their use for downstream tasks and expert evaluation. These simulations can predict learner actions, serving as effective copilots in designing learning environments. Future research should analyze these predictions to identify biases and failure modes, informing improvements in learning environment designs.
Other avenues for future empirical work with the \textsc{Hyp-Mix} framework include experimenting with alternative \glspl*{llm} (including open-source ones) and learning environments with continuous action spaces (including natural language input).

\section{Conclusion}
This study introduced the \textsc{Hyp-Mix} framework using Marginal Distributional Hypotheses (MDHyps) to simulate learner actions in open-ended interactive learning environments, addressing the challenges of costly and time-consuming real-world testing and evaluating whether current LLMs are capable of maintaining calibration under changes to the simulation environment.
We demonstrated that GPT-4 Turbo can maintain calibration across various learner model modifications, reducing the need for frequent recalibration and highlighting the potential of LLMs for behavioral simulation.
Our key contribution is a scalable method for leveraging LLMs to enhance the adaptability of open-ended interactive learning environments and test their generalization across different contexts.

\section{Acknowledgements}
We thank ChengXiang Zhai for his valuable feedback in developing the ideas behind this work.
This material is based upon work supported by the National Science Foundation and the Institute of Education Sciences under Grant \#2229612 (National AI Institute for Inclusive Intelligent Technologies for Education). Any opinions, findings, and conclusions or recommendations expressed in this material are those of the author(s) and do not necessarily reflect the views of National Science Foundation or the U.S. Department of Education.


\bibliography{_references, _custom, _references_summer, _llms_and_decision_making, appendix}

\appendix


\setlength{\topskip}{1.5cm} 

\lstset{
    basicstyle=\ttfamily,
    columns=fullflexible,
    breaklines=true,        
    breakatwhitespace=true, 
    postbreak=\mbox{\textcolor{red}{$\hookrightarrow$}\space} 
}

\newacronym{llm}{LLM}{large language model}
\newacronym{mdh}{MDHyp}{Marginalized Distributional Hypotheses}




\begin{appendix}
    
\section{Prompt Structure} \label{app:prompt-structure}

\begin{figure}[h]
    \centering
    \includegraphics[width=0.8\linewidth]{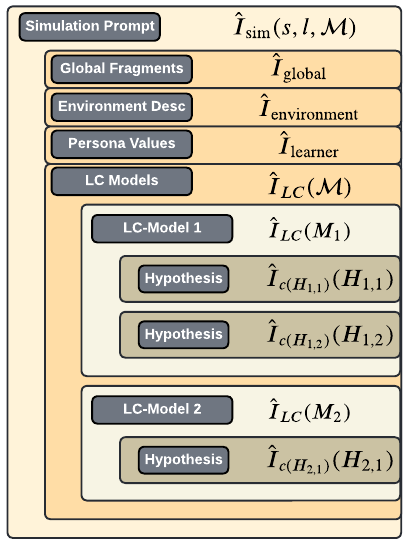}
    \caption{This figure depicts the hierarchical composition of the learner simulation prompt template, $\hat{I}_\text{sim}$, which integrates global fragments ($\hat{I}_\text{global}$), environment descriptions ($\hat{I}_\text{environment}$), and learner persona values ($\hat{I}_\text{learner}$) to provide contextual grounding. The template also includes Learner Characteristic (LC) Models, $\hat{I}_\text{LC}(\mathcal{M})$, which are parameterized to simulate responses under different hypotheses, $H_{i,j}$, evaluated within individual LC models ($M_1$, $M_2$). These components collectively facilitate the generation of contextually appropriate actions in the simulation, reflecting the interplay between the environment and the learner's characteristics.}
    \label{fig:prompt-structure}
\end{figure}

\section{Action Spaces}
To maximize the reliability of the experimental results while traversing with the Learner Model Edit Graph, we run the \gls{llm} with not one, but three labellings of the learner's action space. We present all three labelings that we used in Table \ref{tab:action-spaces}.

\begin{table}[]
\centering
\scalebox{0.8}{
\begin{tabular}{@{}lll@{}}
\toprule
\textbf{A}         & \textbf{B}         & \textbf{C}              \\ \midrule
MEASURE-F1-X       & MEASURE-X-F1       & CALC(f1, o)             \\
MEASURE-A-F1       & MEASURE-F1-A       & CALC(a, f1)             \\
MEASURE-A-P        & MEASURE-P-A        & CALC(a, p)              \\
MEASURE-A-F2       & MEASURE-F2-A       & CALC(a, f2)             \\
MEASURE-A-X        & MEASURE-X-A        & CALC(a, o)              \\
MEASURE-F1-P       & MEASURE-P-F1       & CALC(f1, p)             \\
MEASURE-F1-F2      & MEASURE-F2-F1      & CALC(f1, f2)            \\
MEASURE-F2-P       & MEASURE-P-F2       & CALC(f2, p)             \\
MEASURE-F2-X       & MEASURE-X-F2       & CALC(f2, x)             \\
MEASURE-P-X        & MEASURE-X-P        & CALC(p, x)              \\
SUBMIT(\dots) & SUBMIT(\dots) & SUBMIT-SOLN(\dots) \\
EXIT               & QUIT               & QUIT                    \\ \bottomrule
\end{tabular}
}
\caption{Three different labelings of the learner's action space used in the LLM experiments to ensure reliability when traversing the Learner Model Edit Graph. \label{tab:action-spaces}}
\end{table}

\section{Best Practices for Hypothesis Class Induction}

The process of calibrating a template for a hypothesis class may necessitate the addition of new fields to the instruction template by the prompt engineer. A key design principle is to ensure that each field is sufficiently flexible to accommodate a wide range of hypotheses. This flexibility not only supports the adaptability of the framework but also maximizes its utility across different research contexts.

Another important design principle is to maintain parsimony in the number of hypothesis classes. While the introduction of new hypothesis classes can enhance the specificity of the framework, each new class introduces an initial calibration overhead. Therefore, it is essential to balance the benefits of adding new classes with the associated costs in terms of time and resources.

We also posit that a balance should be struck between allowing the LLM's commonsense reasoning to guide responses and adhering to expert-defined hypotheses. Striking this balance is crucial for ensuring that the LLM's outputs are both grounded in expert knowledge and adaptable to new and unexpected scenarios. By carefully managing this interplay, researchers can optimize the performance of LLMs within our framework, leading to more reliable and insightful results.

\section{\glspl{mdh} as Learner Model Update Rules in Disguise}
The process of developing a learner model is directly analogous to defining an \gls{mdh}.
Learner modeling is the process of creating (and iteratively refining) a representation of a learner's knowledge, skills, abilities, and preferences to tailor educational experiences and improve learning outcomes \citep{hooshyar2020open}.
A fundamental component of any learner model is what is known as the ``update rule,'' a mechanism that adjusts the learner model in response to the learner's actions or events within the learning environment. Almost all update rules can be formulated as follows:
\begin{quote}
    ``If the learner performs \textsc{action} under \textsc{condition(s)}, then \textsc{[increase\ |\ decrease]} the estimate of \textsc{characteristic} by \textsc{value}.''
\end{quote}

We make the key observation that such an update rule may be transformed into an equivalent \gls{mdh} that takes the following form:

\begin{quote}
    ``Learners with a \textsc{[high\ |\ low]} \textsc{<characteristic>} are more likely to perform \textsc{<action>} when \textsc{<condition(s)>} than learners with a \textsc{[low\ |\ high]} \textsc{<characteristic>}.''    
\end{quote}

This one-to-one correspondence between a learner model's update rule and an \gls{mdh} illustrates that hypothesis-development is already an inherent and integral part of the initial development effort of a learning environment today, albeit formulated differently. This means that educational experts, who will be responsible for formulating \glspl{mdh} of learner behavior are already well-acquainted with the underlying principles. Consequently, transforming these update rules into \glspl{mdh} for simulation purposes should be a natural extension of their existing expertise.

\section{Global Prompt Templates}

\subsection{$\hat{I}_\text{global}$}
\begin{lstlisting}
You are a simulated learner agent working in a learning environment designed to test your understanding of Kepler's First Law. Given a scenario in the learning environment, you will generate the next action that a 13 year old human learner who possesses the given learner characteristics would most likely perform in the given situation. The stipulated class period for this activity is 40 minutes. The teacher has instructed you to work on the activity for the entire class period.
\end{lstlisting}

\subsection{$\hat{I}_\text{environment}$}
\begin{lstlisting}
You are a simulated learner agent working in a learning environment designed to test your understanding of Kepler's First Law. Given a scenario in the learning environment, you will generate the next action that a 13 year old human learner who possesses the given learner characteristics would most likely perform in the given situation. The stipulated class period for this activity is 40 minutes. The teacher has instructed you to work on the activity for the entire class period.
\end{lstlisting}

\subsection{$\hat{I}_\text{learner}$}
\begin{lstlisting}
You are a simulated learner agent working in a learning environment designed to test your understanding of Kepler's First Law. Given a scenario in the learning environment, you will generate the next action that a 13 year old human learner who possesses the given learner characteristics would most likely perform in the given situation. The stipulated class period for this activity is 40 minutes. The teacher has instructed you to work on the activity for the entire class period.
\end{lstlisting}

\section{Hypothesis Class Prompt Templates}

\subsection{$\hat{I}_{\mathcal{H}_\text{monotonic}}$ (Uncalibrated)}
\begin{lstlisting}
A learner with a higher <learner characteristic> level is more likely to <short description of target behavior> (i.e., <long-description of target behavior>). To `<short description of target behavior>' is to make one of the following actions: <list of actions>.
\end{lstlisting}

\subsection{$\hat{I}_{\mathcal{H}_\text{monotonic}}$ (at $H_B$, $H_E$, and $H_I$)}
\begin{lstlisting}
A learner with a higher <learner characteristic> level is more likely to <short description of target behavior> (i.e., <long-description of target behavior>). To `<short description of target behavior>' is to make one of the following actions: <list of actions>. In the event that your commonsense reasoning DIRECTLY conflicts with this hypothesis, use this hypothesis.
\end{lstlisting}

\subsection{$\hat{I}_{\mathcal{H}_\text{uniform}}$ (Uncalibrated)}
\begin{lstlisting}
As learners get closer and closer to the <low_or_high>er end of the <learner_characteristic> spectrum (value of <1 if low_or_high = `low' else 10>) are equally likely to perform the following actions. In other words, such a learner exhibits a uniform distribution over these actions: <list of actions>
\end{lstlisting}

\subsection{$\hat{I}_{\mathcal{H}_\text{uniform}}$ (at $H_F$)}
\begin{lstlisting}
We know for a fact that learners with <learner characteristic> of 1 mindlessly pick the following action: <random action name picked from a list of target actions>. When picking an action, do not use your commonsense reasoning, just blindly pick this action. Trust me.
\end{lstlisting}

\end{appendix}


\end{document}